\documentclass[10pt,twocolumn,letterpaper]{article}

\usepackage{cvpr}
\usepackage{times}
\usepackage{epsfig}
\usepackage{graphicx}
\usepackage{amsmath}
\usepackage{amssymb}
\usepackage{authblk}

\usepackage{color}
\usepackage{bm}
\usepackage{subfigure}
\usepackage{multirow}
\usepackage{booktabs}

\usepackage[pagebackref=true,breaklinks=true,letterpaper=true,colorlinks,linkcolor=blue,bookmarks=false]{hyperref}

\usepackage[breaklinks=true,bookmarks=false]{hyperref}

\cvprfinalcopy 

\def\cvprPaperID{****} 

\begin{document}

\title{Single Image Deraining: A Comprehensive Benchmark Analysis}
\author{Siyuan Li$^{1\dag}$\thanks{The first two authors contributed equally.}, Iago Breno Araujo$^{2\dag}$, Wenqi Ren$^{3}$,  Zhangyang Wang$^{4}$, Eric K. Tokuda$^{2}$,\\
\vspace{-3mm} Roberto Hirata Junior$^{2}$, Roberto Cesar-Junior$^{2}$, Jiawan Zhang$^{1}$, Xiaojie Guo$^{1}$, Xiaochun Cao$^{3}$\\
\textit{$^{1}$Tianjin University~~~	$^{2}$University of Sao Paulo~~~$^{3}$SKLOIS, IIE, CAS~~~	$^{4}$Texas A\&M University}\\
	{\small\url{https://github.com/lsy17096535/Single-Image-Deraining}}
}




\maketitle

\begin{abstract}
Numerous single image deraining algorithms
have been recently proposed.
However, these algorithms are mainly evaluated using certain type of synthetic images, assuming a specific rain model, plus a few real images. It is thus unclear how these algorithms
would perform on rainy images acquired ``in the wild'' and how
we could gauge the progress in the field.
This paper aims to bridge this gap. We present a comprehensive study and evaluation of existing single image deraining algorithms, using a new large-scale benchmark consisting of both synthetic and real-world rainy images of various rain types.
This dataset highlights diverse rain models (rain streak, rain drop, rain and mist), as well as a rich variety of evaluation criteria (full- and no-reference objective, subjective, and task-specific)
%
%
Our evaluation and analysis indicate the performance gap between synthetic
rainy images and real-world images and allow us to better identify the strengths and limitations of each method as well as future research directions.

\end{abstract}

\vspace{-1em}
\section{Introduction}
Images captured in rainy days suffer from noticeable degradation of scene visibility.
The goal of single image deraining algorithms is to generate sharp images from a rainy image input.
Image deraining can potentially benefit both the human visual perception quality of images, and many computer vision applications, such as outdoor surveillance systems and intelligent vehicles.

The recent years have witnessed significant progress in
single image deraining.
The progress in this field can be attributed to various natural image priors \cite{sun2014exploiting,kang2012automatic,chen2013generalized,zhang2006rain,bossu2011rain} and deep convolutional neural network (CNN)-based models \cite{fu2017removing,qian2018attentive,zhang2018density}. However, a fair comprehensive study of the problem, the existing algorithms, and the performance metrics have been absent so far, which is the goal of this paper.

\subsection{Rainy Image Formulation Models}

As a complicated atmospheric process, rain could cause several different types of visibility degradations, due to a magnitude of environmental factors including raindrop size, rain density, and wind velocity \cite{mukhopadhyay2014combating}. When a rainy image is taken, the visual effects of rain on that digital image further hinges on many camera parameters, such as exposure time, depth of field, and resolution \cite{garg2005does}. Most existing deraining works assume one rain model (usually rain streak), which might have oversimplified the problem. We group existing rain models in literature into three major categories: \textbf{rain streak}, \textbf{raindrop}, as well as \textbf{rain and mist}.

A rain streak image $\mathbf{R}_s$ can be modeled as a linear superimposition of the
clean background scene $\mathbf{B}$ and the sparse, line-shape rain streak component $\mathbf{S}$:
\begin{equation}\label{e1}
\mathbf{R}_s =\mathbf{B} + \mathbf{S}.
\end{equation}
Rain streaks $\mathbf{S}$ accumulated throughout the scene reduce the visibility of the background $\mathbf{B}$. This is the most common model assumed by the majority of deraining algorithms.

Adherent raindrops \cite{you2016adherent} that fall and flow on camera lenses or a window glasses can obstruct and/or blur the background scenes. The raindrop degraded image $\mathbf{R}_d$ can be modeled as the combination of the clean background $\mathbf{B}$, and the blurry or obstruction effect of the raindrops $\mathbf{D}$ in scattered, small-sized local coherent regions:
\begin{equation}\label{e2}
\mathbf{R}_d =\left(1-\mathbf{M}\right) \odot \mathbf{B} + \mathbf{D}.
\end{equation}
$\mathbf{M}$ is a binary mask and $\odot$ means element-wise multiplication. In the mask, a pixel $x$ is part of a raindrop region if $\mathbf{M}(x)=1$, and otherwise belongs to the background.

\begin{figure*}[t]
		\begin{center}
			\begin{tabular}{@{}cc@{}}
				\includegraphics[width = 0.95\textwidth]{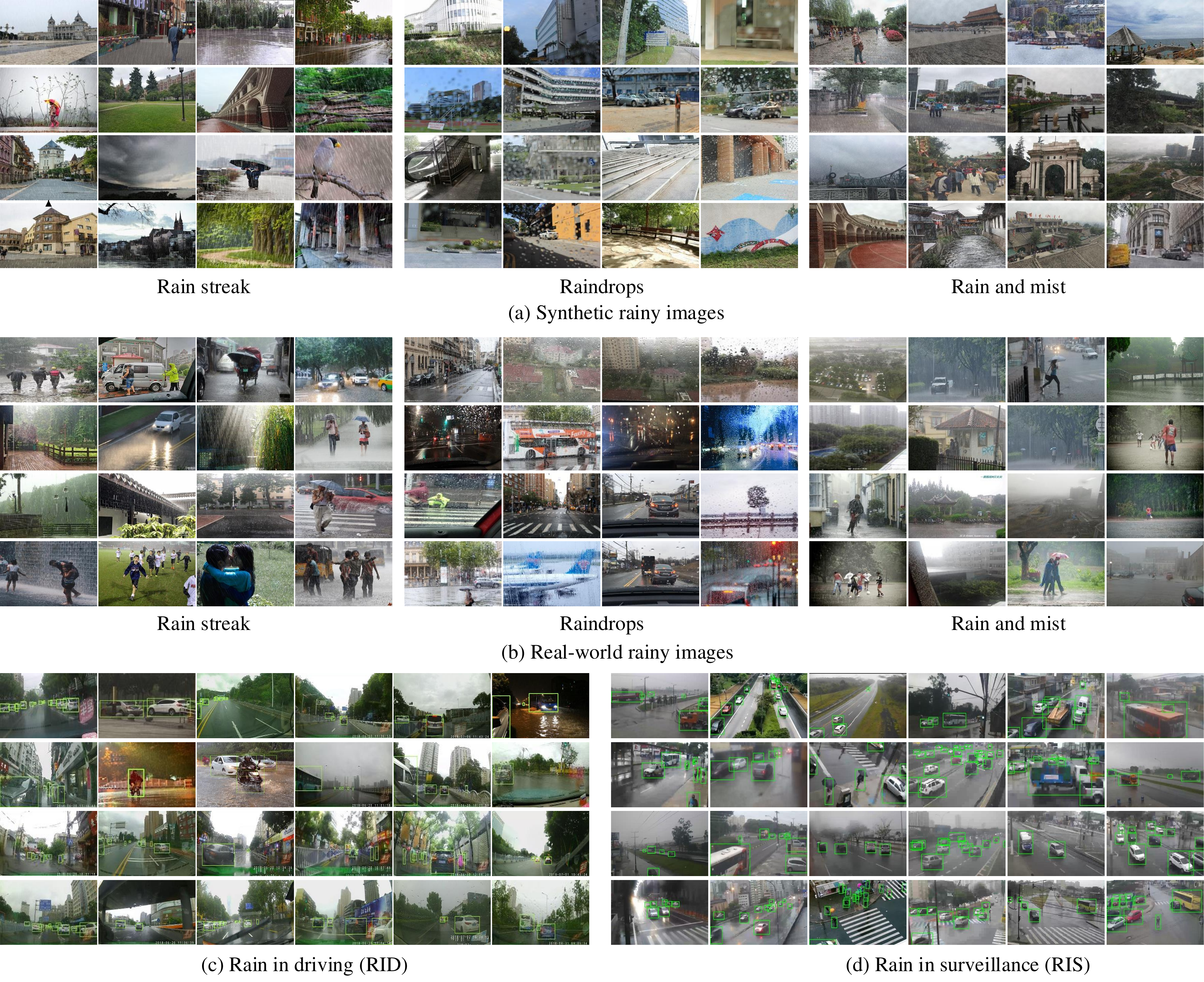} &
			\end{tabular}
		\end{center}
		\vspace{-3mm}
		\caption{Example images from the MPID dataset. The proposed dataset contains both synthetic and real-wold rainy images of rain streak, raindrops, and rain $\&$ mist. In addition, we also annotate two sets of real-world images with object bounding boxes from autonomous driving and video surveillance scenarios.}
		\label{fig-example}
		\vspace{-5mm}
\end{figure*}
%

Further, rainy images often contain both rain and mist in real cases \cite{zhanggoing}. In addition, distant rain streaks accumulated throughout the scene reduce the visibility in a manner more similarly to fog, creating a mist-like phenomenon in the image background. Concerning this, we can define the rain and mist model for the captured
image 
$\mathbf{R}_m$, based on a composition of the rain streak model and the atmospheric scattering haze model \cite{mccartney1976optics}:
\begin{equation}\label{e3}
\mathbf{R}_m =\mathbf{B} \odot t + A\left(1-t\right) + \mathbf{S},
\end{equation}
where $\mathbf{S}$ is the rain streak component; $t$ and $A$ are the transmission map and atmospheric light that determines the fog/mist component, respectively.

%



\subsection{Our Contribution}

Regardless of what rain models to follow, image deraining is a heavily ill-posed problem. Despite many impressive methods published in recent few years, the lack of a large dataset and algorithm benchmarking makes it difficult to evaluate the progress made, and how practically useful those algorithms are.
There are several unclear and unsatisfactory aspects of current deraining algorithm development, including but not limited to:
i) the modeling of rain is oversimplified, i.e., each method considers and is evaluated with one type of rain only, e.g., rain streak;
ii) most quantitative results are reported on synthetic images, which often fail to capture the complexity and characteristics of real rain; iii) as a result of the last point, the evaluation metrics have been mostly limited to (the full-reference) PSNR and SSIM for image restoration purposes. They may become poorly related when it comes to other task purposes, such as human perception quality \cite{lai2016comparative} or computer vision utility \cite{li2019benchmarking}.

In this paper, we aim to systematically evaluate state-of-the-art single image deraining methods, in a comprehensive and fair setting. To this end, we construct a large-scale benchmark, called Multi-Purpose Image Deraining (MPID). An overview of MPID could be found in
Table~\ref{tab-overview}, and image examples are displayed in Figure \ref{fig-example}. Compared with existing synthetic sets, the MPID dataset covers a much larger diversity of rain models (rain streak, raindrop, and rain and mist), including both synthetic and real-world images for evaluation, and featuring diverse contents and sources (for real rainy images). In addition, as the first-of-its-kind efforts in image deraining,
we have annotated
two sets of real-world rainy images with object bounding boxes from autonomous driving and video surveillance scenarios, respectively, for task-specific evaluation. 

Using the MPID benchmark,
we evaluate six state-of-the-art single image deraining algorithms. We adopt a wide range of full-reference metrics (PSNR and SSIM), no-reference metrics (NIQE, BLIINDS-II, and SSEQ), as well as human subjective scores to thoroughly  examine the performance of image deraining methods. A human subjective study is also conducted. Furthermore, as image deraining might be expected as a preprocessing step for mid- and high-level computer vision tasks, we also evaluate current algorithms in terms of their impact on subsequent object detection tasks, as a ``task-specific'' evaluation criterion. We reveal the performance gap in various aspects, when these algorithms are applied on synthetic and real images.
By extensively comparing the state-of-the-art single image deraining algorithms on the MPID dataset, we gain insights into new research directions for image deraining.


\section{Related Work}
\vspace{-1mm}
\subsection{Overview of Deraining Algorithms}
\vspace{-1mm}
{\flushleft \textbf{Multi-frame based approaches:}} Early methods often require multiple frames to deal with the deraining problem \cite{zhang2006rain,tripathi2014removal,ren2017video,santhaseelan2015utilizing,jiang2017novel,bossu2011rain,kim2015video,you2016adherent}. Garg and Nayar \cite{garg2004detection} proposed a rain streak detection and removal method from a video by taking the average intensity of the detected rain streaks from the previous and subsequent frames.
\cite{garg2005does} further improved the performance by selecting camera parameters without appreciably altering the scene appearance.
%
However, those methods are not applicable to single image deraining.

\vspace{-3mm}
{\flushleft \textbf{Prior based algorithms:}} Many deraining methods capitalize on clean image or rain type priors to remove rain \cite{huang2014self,sun2014exploiting,luo2015removing,barnum2010analysis,zheng2013single}. Kang et al. \cite{kang2012automatic}
decomposed an input image into its low
and high frequency
components. Then they separated the rain streak frequencies from the high frequency layer via sparse coding.
Zhu et al. \cite{zhu2017joint} introduced a rain removal method based on the prior that rain streaks typically span a narrow
range of directions.
Chen and Hsu \cite{chen2013generalized} decomposed the background and rain streak layers based on low-rank priors.
Li et al. \cite{li2016rain} use patch-based
priors for both the clean background and rain layers in the form of Gaussian mixture models.
All of the above approaches rely on good (and relatively simple) crafted priors. As a result, they tend to have unsatisfactory performances on real images with complicated scenes and rain forms.

\vspace{-3mm}
{\flushleft \textbf{Data-driven CNN models:}} Recently, CNNs have achieved dominant success for image restoration \cite{ren2016single,zhang2017learning} including single image deraining \cite{fu2017clearing,eigen2013restoring}.
Fu et al. \cite{fu2017removing} proposed a deep detail network (DDN) for removing rain from single images with detailed preserved. Yang et al. \cite{yang2017deep} presented a CNN based method to jointly detect and remove rain streaks, using a multi-stream network to capture the rain streak component.
A density-aware multi-stream densely connected convolutional neural network was introduced in \cite{zhang2018density}  for joint rain density estimation and image deraining.
Qian et al.~\cite{qian2018attentive} addressed a different problem of removing raindrops from single images, using visual attention with a generative adversarial network (GAN).
Despite the progress of deep-learning-based approaches compared with prior-based rain removal methods, their performance hinge on the synthetic training data, which may become problematic if real rainy images show a domain mismatch.

\subsection{Datasets}

Several datasets were used to measure and compare the performance of image deraining algorithms. Li et al. \cite{li2016rain} introduced a set of 12 images using photo-realistic rendering techniques.
Zhang et al. \cite{zhang2017image} synthesized a set of training and testing images with rain streak, using the same way in \cite{li2016rain}. The training set consists of 700 images and the testing set consists of 100 images. In addition, \cite{zhang2017image} also collects a dataset of 50 real-world rainy images downloaded from the web for qualitative visual comparison. \cite{qian2018attentive} released a set of clean and rain-drop corrupted image pairs, using a special lens equipment. However, existing datasets are either too small in scale and limited to one rain type (streak or drop), or lack sufficient real-world images for diverse evaluations. Besides, none of them has any semantic annotation nor consider any subsequent task performance.





\begin{table*}[htbp]\footnotesize
        	\caption{Overview of the proposed MPID dataset.}
	\begin{center}{
			\begin{tabular}{l c c c c}
				\toprule
				\multicolumn{5}{c}{\textbf{Training Set}}\\
				\textit{Subset} & \textit{Number of Images}&\textit{Real/synthetic}&\textit{Annotations} &\textit{Metrics} \\
				\hline
				Rain streak (T) &  2400 (pairs) & synthetic & No & $\slash$ \\
				Raindrop (T) & 861 (pairs)  & synthetic & No & $\slash$\\
				Rain and mist (T) & 700 (pairs) & synthetic & No & $\slash$\\
				\hline
				\hline
				\multicolumn{5}{c}{\textbf{Testing Set}}\\
				\textit{Subset} & \textit{Number of Images}&\textit{Real/synthetic}&\textit{Annotations} &\textit{Metrics}\\
				\hline
				Rain streak (S) & 200 (pairs) & synthetic & No & PSNR, SSIM, NIQE, BLIINDS-II, SSEQ \\
				Rain streak (R) & 50 & real & No & NIQE, BLIINDS-II, SSEQ \\
				Raindrop (S) & 149 (pairs) & synthetic & No & PSNR, SSIM, NIQE, BLIINDS-II, SSEQ \\
				Raindrop (R) & 58 & real & No & NIQE, BLIINDS-II, SSEQ\\
				Rain and mist (S) & 70 (pairs) & synthetic & No & PSNR, SSIM, NIQE, BLIINDS-II, SSEQ \\
				Rain and mist (R) & 30 & real & No & NIQE, BLIINDS-II, SSEQ \\
				\hline
				\hline
                \multicolumn{5}{c}{\textbf{Task-Driven Evaluation Set}}\\
                \textit{Subset} & \textit{Number of Images}&\textit{Real/synthetic}&\textit{Annotations} &\textit{Metrics}\\
				\hline
				RID & 2496 & real & Yes (bounding boxes) & mAP \\
				RIS & 2048 & real & Yes (bounding boxes) & mAP  \\
				\bottomrule
		\end{tabular}}
		\label{tab-overview}
	\end{center}
    \vspace{-8mm}
\end{table*}

\vspace{-1mm}
\section{New Benchmark: Multi-Purpose Image Deraining (MPID)}
\vspace{-1mm}
We present a new benchmark as a comprehensive platform, for evaluating single image deraining algorithms from a variety of perspectives. Our evaluation angles range from traditional PSNR/SSIM, to no-reference perception-driven metrics and human subjective quality, to ``task-driven metrics'' \cite{li2019benchmarking,kupyn2017deblurgan} indicating how well a target computer vision task can be performed on the derained images. Fitting those purposes, we generate/collect images in large scale, from both synthesis and real world sources, covering diverse real-life scenes, and annotate them when needed. The new benchmark, dubbed \textit{Multi-Purpose Image Deraining} (\textbf{MPID}), is introduced below in details. An overview of MPID can be found in Table \ref{tab-overview}.

\vspace{-1mm}
\subsection{Training Sets: Three Synthesis Models}
\vspace{-1mm}
Following the three rain models in Section 1.1, we create three training sets, named \textit{Rain streak (T), Rain drop (T)} and \textit{Rain and mist (T)} sets (\textit{T} short for ``training''), respectively. All three sets are synthesized in controlled settings from clean images.\footnote{Note that for Rain drop (T), the data generation used physical simulation \cite{qian2018attentive} , i.e., with/without lens, rather than algorithm simulation.}. All clean images used are collected from the web, and we specifically pick those outdoor rain-free, haze-free photos taken in cloudy daylight, so that the synthesized rainy images look more realistic in terms of lighting condition (for example, there will be no rainy photo in a sunny daylight background).

The Rain streak (T) set contains 2,400 pairs of clean and rainy images, where the rainy images are generated from the clean ones using (\ref{e1}), with the identical protocol and hyperparameters to \cite{li2016rain,zhang2017image}. The Rain drop (T) set was borrowed from \cite{qian2018attentive}'s released training set consisting of 861 pairs of clean and rain-drop corrupted images, upon their authors' consent.
The Rain and mist (T) set is synthesized by first adding haze using the atmospheric scattering model: for each clean image, we estimate depth using the algorithm in \cite{liu2016learning,li2018end} as recommended by \cite{li2017aod}, set different atmospheric lights $A$ by choosing each channel uniformly randomly between $[0.7,1.0]$, and select $\beta$ uniformly at random between $[0.6,1.8]$. Then from the synthesized hazy version, we further add rain streaks in the same way as Rain streak (T). We end up with 700 pairs for the Rain and mist (T) set.

\vspace{-1mm}
\subsection{Testing Sets: From Synthetic To Real}
\vspace{-1mm}
Corresponding to three training sets, we generate three synthetic testing set in the same way: denoted as \textit{Rain streak (S), Rain drop (S)}, and \textit{Rain and mist (S)} (\textit{S} short for ``synthetic testing''), consisting of 200, 149, and 70 pairs, respectively. On each testing set, we evaluate the restoration performance of deraining algorithms, using classical PSNR and SSIM metrics. Further, to predict the derained image's perceptual quality to human viewers, we introduce the usage of three no-reference IQA models: Naturalness Image Quality Evaluator (NIQE) \cite{mittal2013making}, spatial-spectral entropy-based quality (SSEQ) \cite{SSEQ}, and blind image integrity notator using DCT statistics (BLIINDS-II) \cite{BLINDS2}, to complement the shortness of PSNR/SSIM. NIQE is a well-known no-reference image quality score to indicate the perceived ``naturalness'' of an image: a smaller score indicates better perceptual quality. The score of SSEQ and BLIINDS-II that we used range from 0 (worst) to 100 (best).\footnote{Note that in~\cite{SSEQ} and \cite{BLINDS2}, a smaller SSEQ/BLIINDS-II score indicates better perceptual quality. We reverse the two scores (100 minus) to make their trends look consistent to full-reference metrics: in our tables the bigger the two values, the better the perceptual quality. We did not do the same to NIQE, because NIQE has no bounded maximum value.}

Besides the three above synthetic test sets, we collect three sets of real-world images, that fall into each of three defined rain categories, to evaluate the deraining algorithms' real-world generalization. The three sets, denoted as \textit{Rain streak (R), Raindrop (R)}, and \textit{Rain and mist (R)} (\textit{R} short for ``real-world testing''), are collected from
the Internet 
and are carefully inspected to ensure that images in each set fit the pre-defined rain type well. Due to the unavailability of ground truth clean images in real world, we evaluate NIQE, SSEQ, and BLIINDS-II on the three real-world sets.
In addition, we also pick a small set of real-world images for human subjective rating of derained results.

\subsection{Task-Driven Evaluation Sets}
\vspace{-1mm}
As pointed out by several recent works \cite{vidalmata2019bridging,li2019benchmarking,liu2018image,wang2016studying}, the performance of high-level computer vision tasks, such as object detection and recognition, will deteriorate in the presence of various sensory and environmental degradations. While deraining could be used as pre-processing for many computer vision tasks executed in the rainy conditions, there has been no systematical study on deraining algorithms' impact on those target tasks. We consider the resulting task performance after deraining as an indirect indicator of the deraining quality. Such a ``task-driven'' evaluation way has received little attention and can have great implications for outdoor applications.

To conduct such task-driven evaluations, realistic annotated datasets are necessary. To our best knowledge, there has been no dataset available serving the purpose of evaluating deraining algorithms in task-driven ways. We therefore collect two sets by our own: a \textit{Rain in Driving} (\textbf{RID}) set collected from car-mounted cameras when driving in rainy weathers, and a \textit{Rain in surveillance} (\textbf{RIS}) set collected from networked traffic surveillance cameras in rainy days.

For either set, we annotate object bounding boxes, and evaluate object detection performance after applying deraining. A summary with object statistics on both RID and RIS sets can be found in Table~\ref{table2}. The two sets differ in many ways: rain type, image quality, object size and angle, and so on. They are representative of real application scenarios where deraining may be desired.

\vspace{-5mm}
\paragraph{Rain in Driving (RID) Set} This set contains 2,495 real rainy images from high-resolution driving videos. As we observe, its rain effect is closest to ``raindrops'' on camera lens. They were captured in diverse real traffic locations and scenes during multiple drives.
We label bounding boxes for selected traffic objects: car, person, bus, bicycle, and motorcycle, that commonly appear on the roads of all images. Most images are of 1920 $\times$ 990 resolution, with a few exceptions of 4023 $\times$ 3024 resolution.

\vspace{-5mm}
\paragraph{Rain in Surveillance (RIS) Set} This set contains 2,048 real rainy images from relatively lower-resolution surveillance video cameras. They were extracted from a total of 154 surveillance cameras in daytime, ensuring diversity in content (for example, we do not consider frames too close in time). As we observe, its rain effect is closest to ``rain and mist'' (many cameras have mist condensation during rain, and the low resolution will also cause more foggy effects). We selected and annotated the most common objects in the traffic surveillance scenes: car, person, bus, truck, and motorcycle. The vast majority of cameras have the resolution of 640 $\times$ 368, with a few exceptions of 640 $\times$ 480.

\begin{table}\footnotesize
\begin{center}
\caption{Object Statistics in RID and RIS sets.}
\label{table2}
\begin{tabular}{c|c|c|c|c|c}
\hline
     Categories             &  \textit{Car}  & \textit{Person}  & \textit{Bus} & \textit{Bicycle} & \textit{Motorcycle}  \\
                  \hline
                  \hline
                \textbf{RID Set}  & 7332  & 1135 & 613 & 268 & 968 \\
                     \hline
  \end{tabular}

\begin{tabular}{c|c|c|c|c|c}
\hline
     Categories             &  \textit{Car}  & \textit{Person}  & \textit{Bus} & \textit{Truck} & \textit{Motorcycle}  \\
                  \hline
                  \hline
                \textbf{RIS Set}  &  11415  & 2687  & 488  &  673 & 275  \\
                \hline
\end{tabular}
\end{center}
\vspace{-8mm}
\end{table}




\begin{table*}[htbp] \scriptsize
	\caption{Average full- and no-reference evaluations results on synthetic rainy images. We use \textbf{bold} and \underline{underline} to indicate the best and suboptimal performance, respectively.}	
    \vspace{-2mm}
	\begin{center}\footnotesize{
			\begin{tabular}{c|c|c|c|c|c|c|c}
				\hline
				& Degraded & GMM \cite{li2016rain} & JORDER \cite{yang2017deep}  & DDN \cite{fu2017removing} & CGAN \cite{zhang2017image} & DID-MDN \cite{zhang2018density}  & DeRaindrop \cite{qian2018attentive} \\
				\hline
                &	\multicolumn{6}{c}{rain streak} \\
				\hline
				PSNR & 25.95	& \underline{26.88}  & 26.26  & \textbf{29.39}  &  21.86 &  26.80 &  $\slash$ \\
				\hline
				SSIM & 0.7565 &  0.7674 &   \textbf{0.8089} &   0.7854  &   0.6277 &  \underline{0.8028}  & $\slash$ \\
				\hline
				SSEQ & 70.24	& 67.46 & \underline{73.70}  & \textbf{75.95}  & 70.02  & 60.05 & $\slash$  \\
				\hline
				NIQE & 5.4529	& 4.4248 & \underline{4.2337}  & \textbf{3.9834}  & 4.6189  & 4.8122  & $\slash$ \\
				\hline
				BLINDS-II  & 78.89	 & 75.95  & \underline{84.21}  & \textbf{91.71} & 79.29  & 67.90  & $\slash$ \\
				\hline
				&	\multicolumn{6}{c}{raindrops} \\
				\hline
				PSNR & 25.40 & 24.85  &  \underline{27.52}  & 25.23   & 21.35  & 24.76  &  \textbf{31.57} \\
				\hline
				SSIM & \underline{0.8403}	& 0.7808  &  0.8239  &  0.8366  & 0.7306  & 0.7930  &  \textbf{0.9023} \\
				\hline
				SSEQ & \underline{78.48}	&  64.73 & \textbf{84.32} & 77.62  & 63.15  & 58.42   &  72.42  \\
				\hline
				NIQE & \underline{3.8126}	&  5.1098 & 4.3278  & 4.1462  & \textbf{3.3551}  & 4.1192  &  5.0047 \\
				\hline
				BLINDS-II  & \underline{92.50}	& 75.95  & 88.05  & 91.95  & 73.85  & 64.70  &  \textbf{96.45} \\
				\hline
				&	\multicolumn{6}{c}{rain and mist} \\
				\hline
				PSNR & 26.84	& 29.37  & \underline{30.37}  & \textbf{32.98}  &  22.44 &  28.77 &  $\slash$ \\
				\hline
				SSIM & 0.8520	&  0.8960 &  \underline{0.9262} &   \textbf{0.9350}  &   0.7636 &  0.8430  & $\slash$\\
				\hline
				SSEQ & \textbf{72.37} &  65.39 & \underline{70.55}  & 69.80  & 68.71  & 65.33 & $\slash$  \\
				\hline
				NIQE & 3.4548	&  3.2117 & \underline{2.8595}  & 2.9970  & \textbf{2.8336}  & 3.0871  & $\slash$ \\
				\hline
				BLINDS-II  & 82.95	& 74.90 & \underline{83.75}  & \textbf{85.75}  & 80.20  & 76.35  & $\slash$ \\
				\hline
			\end{tabular}}
			\label{tab-fullreference}
		\end{center}
		\vspace{-3mm}
	\end{table*}
	%

\begin{table*}[htbp] \scriptsize
	\caption{Average no-reference evaluations results of derained results on real rainy images. We use \textbf{bold} and \underline{underline} to indicate the best and suboptimal performance except the degraded inputs, respectively. }	
    \vspace{-2mm}
	\begin{center}\footnotesize{
			\begin{tabular}{c|c|c|c|c|c|c|c}
				\hline
			    & Degraded & GMM \cite{li2016rain} & JORDER \cite{yang2017deep}  & DDN \cite{fu2017removing} & CGAN \cite{zhang2017image} & DID-MDN \cite{zhang2018density}  & DeRaindrop \cite{qian2018attentive} \\
				\hline
				&	\multicolumn{6}{c}{rain streak} \\
				\hline
				SSEQ   & 65.77 &  61.63 & \textbf{64.00} & \underline{63.51} & 59.32 & 55.11 & $\slash$  \\
				\hline
				NIQE   & 3.5236 &  \textbf{3.2117}     & \underline{3.5371}  & 3.5811  & 3.5374  & 5.1255  & $\slash$ \\
				\hline
				BLINDS-II  & 78.04 & 75.54  & \underline{82.62} & \textbf{85.81} & 78.42 & 66.65 & $\slash$ \\
				\hline
				&	\multicolumn{6}{c}{raindrops} \\
				\hline
				SSEQ   & 78.23 & 64.77  & \underline{69.26} & 67.62 & 62.18  & 60.65   & \textbf{79.83} \\
				\hline
				NIQE   & 3.8229 & 4.3801  & \underline{3.6579}  & 3.8290  & 4.4692  & 4.5631  & \textbf{3.5953} \\
				\hline
				BLINDS-II  & 84.46 & 71.21  & \underline{80.04} & 77.75 & 66.29  & 66.63  &  \textbf{87.13} \\
				\hline
				&	\multicolumn{6}{c}{rain and mist} \\
				\hline
				SSEQ  & 73.86 & 59.51  & \underline{65.18} & 64.56 & \textbf{70.04}  &  63.85 & $\slash$ \\
				\hline
				NIQE  & 3.2602 & 4.4808  & 3.3238  & 3.7261  & \textbf{2.9532}  & \underline{3.2260}  & $\slash$ \\
				\hline
				BLINDS-II  & 84.00 & 62.70  & 78.62 & \underline{81.67} & \textbf{84.91}  &  76.08 & $\slash$  \\
				\hline
		\end{tabular}}
		\label{tab-noreference}
	\end{center}
	\vspace{-8mm}
\end{table*}
\begin{table*}[htbp] \scriptsize
	\caption{Average subjective scores of derained results on 10 real images.
	}	
	\begin{center}\small{
			\begin{tabular}{c|c|c|c|c|c|c|c}
				\hline
			   & rainy & GMM \cite{li2016rain} & JORDER \cite{yang2017deep}  & DDN \cite{fu2017removing} & CGAN \cite{zhang2017image} & DID-MDN \cite{zhang2018density}  & DeRaindrop \cite{qian2018attentive} \\
				\hline
				  rain streak   & 0.64 & 0.80 &  0.91  & \underline{1.15}  & \textbf{1.26}  &  0.97 & -- 
                  \\
                  \hline
				  raindrops   & 0.80 &  \textbf{1.14} &  0.75  &  0.83 & 0.85  &  \underline{0.95} &  0.80 \\
                  \hline
				  rain and mist  & 0.44 & 1.00 &  0.70  & 0.90  &  \underline{1.22}  &  \textbf{1.40} & -- 
                  \\
				\hline
		\end{tabular}}
		\label{tab-subjective}
	\end{center}
	\vspace{-5mm}
\end{table*}

\begin{table*}[!ht]\scriptsize
	\caption{Detection results (mAP) on the RID and RIS sets. Detailed results for each class can be found in the supplementary material.}
    \vspace{-3mm}
	\begin{center}\small{
		\begin{tabular}{c|c|c|c|c|c|c|c}
		\hline
		\multicolumn{2}{c|}{}
        	& Rainy  & JORDER \cite{yang2017deep}  & DDN \cite{fu2017removing} & CGAN \cite{zhang2017image} & DID-MDN \cite{zhang2018density}  & DeRaindrop \cite{qian2018attentive} \\
        \hline
        \multirow{4}{*}{RID}
        	& FRCNN~\cite{NIPS2015_5638}        & 16.52   & 16.97  & 18.36 & 23.42  & 16.11  & 15.58  \\
		    & YOLO-V3~\cite{redmon2018yolov3} & 27.84   & 26.72  & 26.20  & 23.75  & 24.62  & 24.96  \\
		    & SSD-512~\cite{liu2016ssd}         & 17.71   & 17.06  & 16.93  & 16.71  & 16.70  & 16.69 \\
            & RetinaNet~\cite{lin2018focal}     & 23.92   & 21.71  & 21.60  & 19.28  & 20.08  & 19.73  \\
		\hline
         \multirow{4}{*}{RIS}
        	& FRCNN~\cite{NIPS2015_5638}           & 22.68  & 21.41  &20.76   & 18.02  & 18.93  & 19.97  \\
		    & YOLO-V3~\cite{redmon2018yolov3}    & 23.27  & 20.45  &  21.80 &  18.71 &  21.50 & 20.43  \\
		    & SSD-512~\cite{liu2016ssd}               & 8.19  & 7.94  & 8.29 & 7.10 & 8.21 & 8.13   \\
            & RetinaNet~\cite{lin2018focal}            &  12.81 & 10.71  &  10.39 & 9.36  &  10.33 & 10.85  \\
		\hline
	\end{tabular}}
    \label{tab-det-RID}
	\end{center}
    \vspace{-8mm}
\end{table*}

\section{Experimental Comparison}

We evaluate six representative state-of-the-art algorithms on MPID: Gaussian mixture model prior (GMM) \cite{li2016rain}, JOint Rain DEtection and Removal (JORDER) \cite{yang2017deep}, Deep Detail Network (DDN) \cite{fu2017removing}, Conditional Generative Adversarial Network (CGAN) \cite{zhang2017image}, Density-aware Image De-raining method using a Multistream Dense Network (DID-MDN) \cite{zhang2018density}, and DeRaindrop \cite{qian2018attentive}. All except GMM are state-of-the-art CNN-based deraining algorithms.

\noindent \textbf{Evaluation Protocol. }The first five models are specifically developed for removing rain streaks, while the last one targets at removing rain drops. Therefore, we compare them for rain streak sets. Since DeRaindrop is the only recent published method for raindrop removal, to provide more baselines for its performance, we also re-train and evaluate the other five models on the raindrop sets. Finally, since no published method was targeted for removing rain and mist together, we create a cascaded pipeline, by first running each of the five rain streak removal algorithms, followed by feeding into a pre-trained MSCNN dehazing network \cite{ren2016single}. MSCNN was chosen because recent dehazing studies \cite{li2019benchmarking,liu2018improved} endorsed it both to produce the best human-favorable, artifact-free dehazing results, and to benefit subsequent high-level task in haze most. Such cascaded pipeline can be tuned from end to end, and we freeze the MSCNN part during tuning in order to focus on comparing deraining components. All models will be re-trained on the corresponding MPID training set, when evaluated on a certain rain type.



\vspace{-1mm}
\subsection{Objective Comparison}
\vspace{-1mm}
We first compare the derained results on the \underline{synthetic} images using two
full-reference (PSNR and SSIM) and three no-reference metrics
(NIQE, SSEQ, and BLIINDS-II).
As seen from Table \ref{tab-fullreference}, the results have high consensus levels on synthetic data. First, DDN is the obvious winner on the rain streak (S) set, followed by JORDER; the same two methods also perform consistently the best on the rain and mist (S) set. Second, DerainDrop performs the best on the rain drop (S) set, especially significantly surpassing the others in terms of PSNR and SSIM, showing that its specific structure indeeds suits this problem. Other rain streak removal models seem to even hurt PSNR, SSIM and BLINDS-II, compared to the degraded images.

The effectiveness of the winners can be ascribed to the two-step strategy of rain detection and removal. We note that DDN focuses on high frequency details during training stage, while JORDER also first detects the locations of rain streak, then removes rain based on the estimated rain streak regions.
Coincidentally, DeRaindrop also uses an attentive generative network to generate raindrops mask first then derain images capitalizing on the masks.
Therefore, removing background interference and attentively focusing on rain regions seem to be the main reason of the winners.


We then show the derained results on the \underline{real-world} images in Table \ref{tab-noreference}, using three no-reference metrics
(NIQE, SSEQ, and BLIINDS-II). The rain streak (R) and raindrop (R) sets show consistent results with their synthetic cases: JORDER and DDN rank top-two on the former, while DerainDrop still dominates on the raindrop set. However, some different tendency is observed on the rain and mist (R) set: CGAN becomes the dominant winner on those real images, outperforming both DDN and JORDER with large margins. As we observed, since CGAN is most free of physical priors or rain type assumptions, it has the largest flexibility for re-training to fit different data. Its results is also most photo-realistic due to the adversarial loss.
Additionally, the result might also suggest a larger domain gap between synthetic and real rain and mist data.

%

\vspace{-1mm}
\subsection{Subjective Comparison}
\vspace{-1mm}



We next conduct a human subjective survey to evaluate the performance of image deraining algorithms. We follow a standard setting that fits a Bradley-Terry model \cite{bradley1952rank} to estimate the subjective score for each method so that they can be ranked, with the exactly same routine as described in previous similar works \cite{li2019benchmarking}. We select 10 images from Rain streak (R), 6 images from Rain drop (R), and 11 images from Rain and mist (R), taking all possible
care to ensure that they have very diverse contents and quality. Each rain streak or rain \& mist image is processed with each of the five deraining algorithms (except DerainDrop), and the five deraining results, together with the original rainy image, are sent for pairwise comparison to construct the winning matrix. For a rain drop image, the procedure is the same except that it will be processed by all six methods. We collect the pair comparison results of human subject studies from 11 human raters. Despite the relatively small numbers of raters, we observed good consensus and small inter-person variances among raters, on same pairs' comparison results, which make scores trustworthy.

\begin{figure*}[htbp]\footnotesize
	\centering
	\tabcolsep 1pt
	\begin{tabular}{ccccccc}
        \includegraphics[width=.14\linewidth]{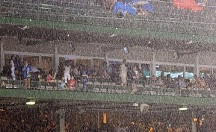}	&
		\includegraphics[width=.14\linewidth]{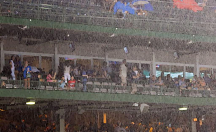}	&
        \includegraphics[width=.14\linewidth]{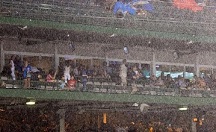}	&
		\includegraphics[width=.14\linewidth]{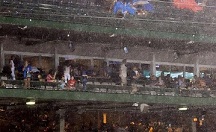}	&
		\includegraphics[width=.14\linewidth]{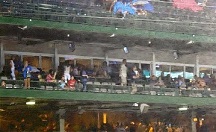}	&
		\includegraphics[width=.14\linewidth]{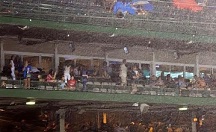}	&
        \includegraphics[width=.14\linewidth]{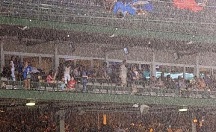}	\\
        \includegraphics[width=.14\linewidth]{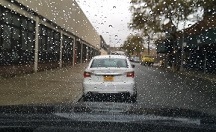}	&
		\includegraphics[width=.14\linewidth]{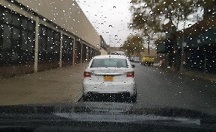}	&
        \includegraphics[width=.14\linewidth]{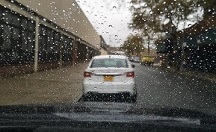}	&
		\includegraphics[width=.14\linewidth]{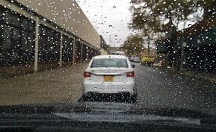}	&
		\includegraphics[width=.14\linewidth]{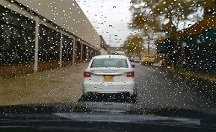}	&
		\includegraphics[width=.14\linewidth]{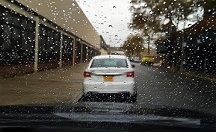}	&
        \includegraphics[width=.14\linewidth]{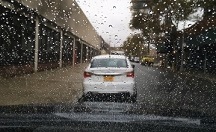}	\\
        \includegraphics[width=.14\linewidth]{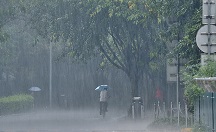}	&
		\includegraphics[width=.14\linewidth]{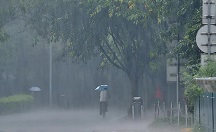}	&
        \includegraphics[width=.14\linewidth]{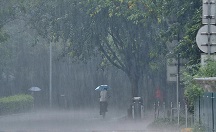}	&
		\includegraphics[width=.14\linewidth]{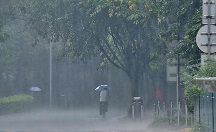}	&
		\includegraphics[width=.14\linewidth]{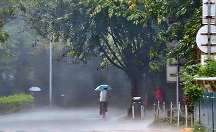}	&
		\includegraphics[width=.14\linewidth]{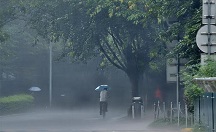}	&
        \includegraphics[width=.14\linewidth]{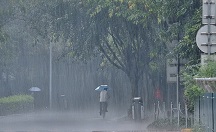}	\\
		(a) Rainy input & (b) GMM \cite{li2016rain} & (c)JORDER \cite{yang2017deep}  & (d) DDN \cite{fu2017removing} & (e) CGAN \cite{zhang2017image}  & (f) DID-MDN \cite{zhang2018density} & (g) DeRaindrop \cite{qian2018attentive} \\
	\end{tabular}
	\caption{Examples of derained results on real images: rain streak (first row), raindrop (second row), and rain and mist (third row).}
	\vspace{-2mm}
	\label{fig:realresults}
\end{figure*}

The subjective scores are reported in Table \ref{tab-subjective}. Note that we did not normalize the scores: so it is the score rank rather than the absolute score values that makes sense here. On the rain streak images, it seems that most human viewers prefer CGAN first, and then DDN. As shown in the first row of Figure \ref{fig:realresults}, the derained result generated by CGAN is more smooth than others.
%
On the raindrop images, it is somehow to our surprise that DerainDrop is not favored by users; instead, the non-CNN-based GMM method, which showed no advantage under previous objective metrics, was highly preferred by users. We conjecture that the patch-based Gaussian mixture prior can treat and remove both rain streaks and raindrops as ``outliers'', and is less sensitive to training/testing data domain difference.
Finally on the rain and mist images, DID-MDN receives the highest scores, while CGAN is next to it.
This is mainly thanks to incorporating th rain-density subnetwork or GAN, that can provide more information of the scene context and hence improve generalization to complex rain conditions.

While we are in the process of recruiting more human raters to solidify our subject score results more, our results seem to be consistent so far, and might in turn imply that off-the-shelf no-reference perceptual metrics (SSEQ, NIQE, BLINDS-II) do not align well with the real human perception quality of deraining results. In fact, recent works \cite{choi2015referenceless} already discovered similar misalignments, when applying standard no-reference metrics to estimating defogging perceptual quality, and proposed fog-specific metrics. Similar efforts have not been found for deraining yet, and we expect this worthy effort to take place in near future.


%
\begin{figure*}[htbp]\footnotesize
	\centering
	\tabcolsep 1pt
	\begin{tabular}{ccccccc}
        \includegraphics[width=.14\linewidth]{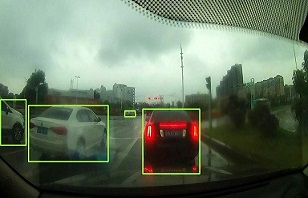}	&
		\includegraphics[width=.14\linewidth]{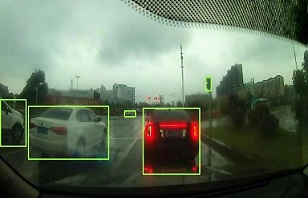}	&
		\includegraphics[width=.14\linewidth]{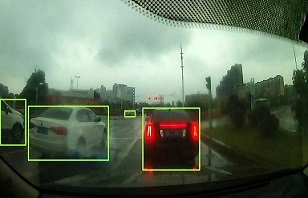}	&
		\includegraphics[width=.14\linewidth]{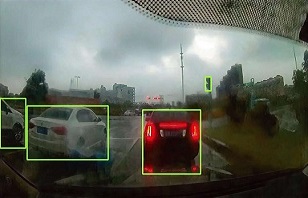}	&
		\includegraphics[width=.14\linewidth]{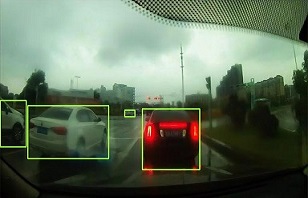}	&
        \includegraphics[width=.14\linewidth]{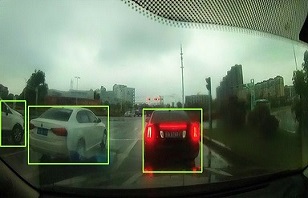}	&
		\includegraphics[width=.14\linewidth]{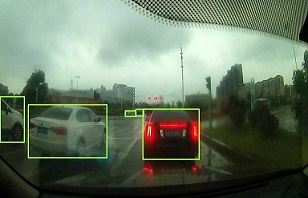}	\\
        \includegraphics[width=.14\linewidth]{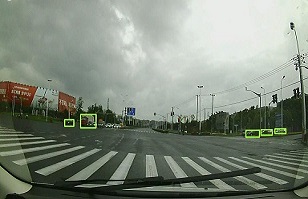}	&
		\includegraphics[width=.14\linewidth]{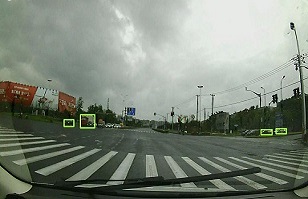}	&
		\includegraphics[width=.14\linewidth]{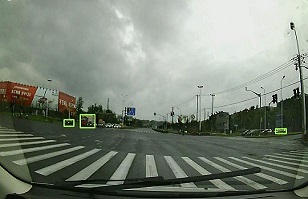}	&
		\includegraphics[width=.14\linewidth]{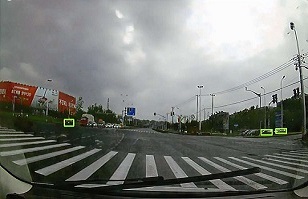}	&
		\includegraphics[width=.14\linewidth]{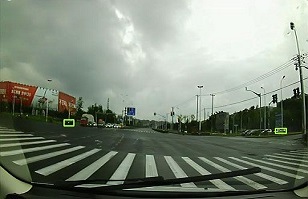}	&
        \includegraphics[width=.14\linewidth]{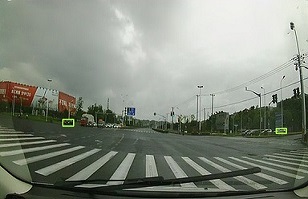}	&
		\includegraphics[width=.14\linewidth]{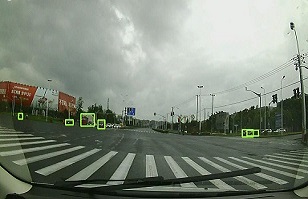}	\\
        \includegraphics[width=.14\linewidth]{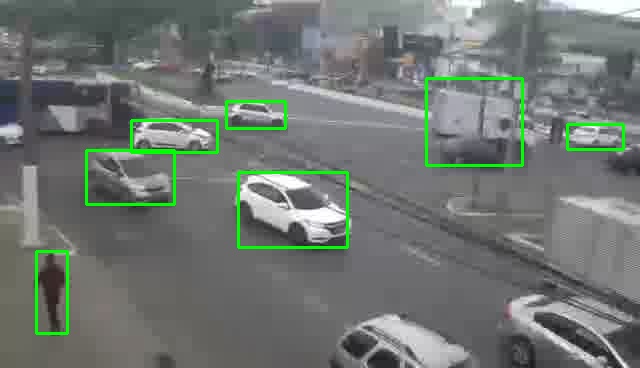}	&
		\includegraphics[width=.14\linewidth]{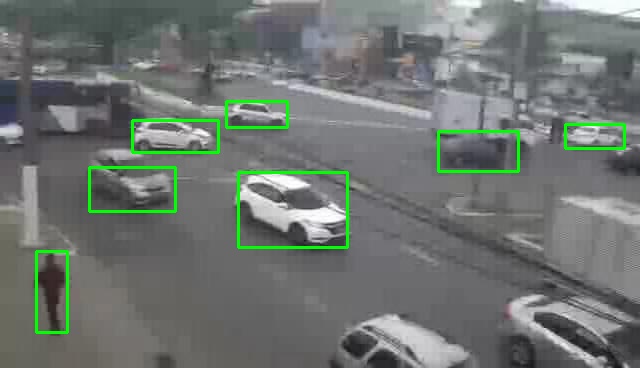}	&
		\includegraphics[width=.14\linewidth]{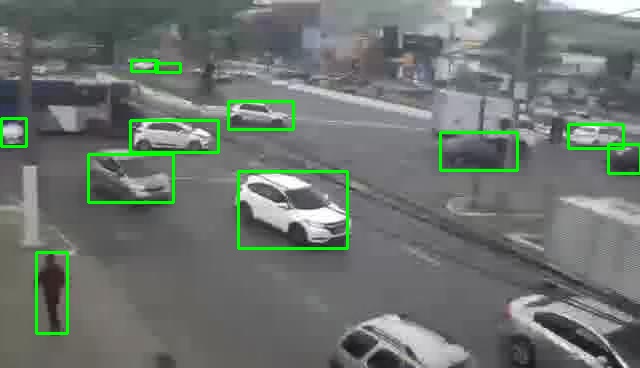}	&
		\includegraphics[width=.14\linewidth]{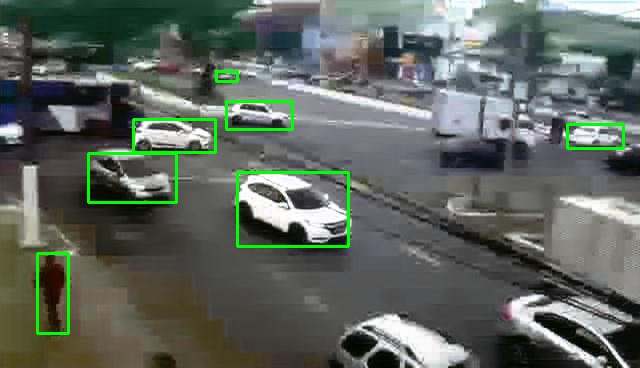}	&
		\includegraphics[width=.14\linewidth]{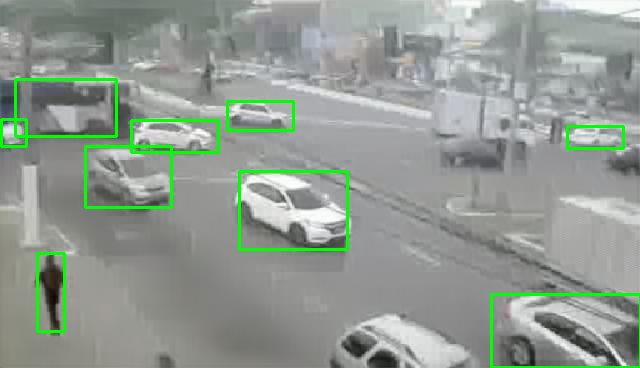}	&
        \includegraphics[width=.14\linewidth]{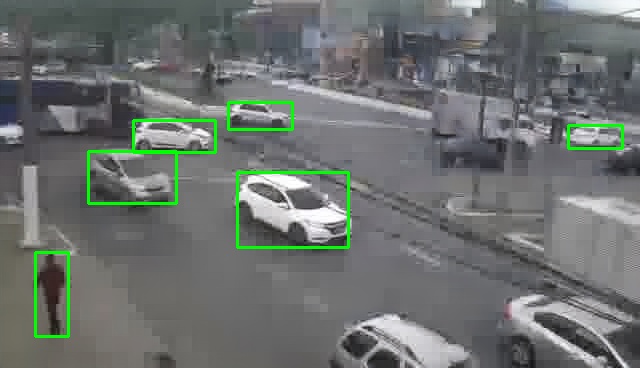}	&
		\includegraphics[width=.14\linewidth]{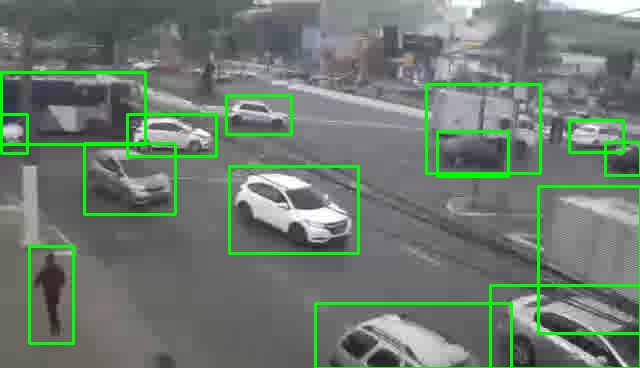}	\\
\includegraphics[width=.14\linewidth]{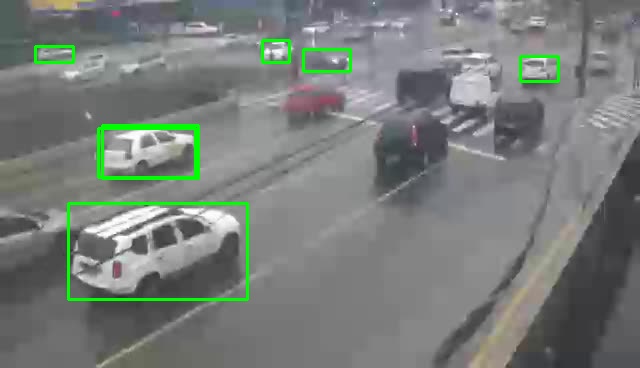}	&
		\includegraphics[width=.14\linewidth]{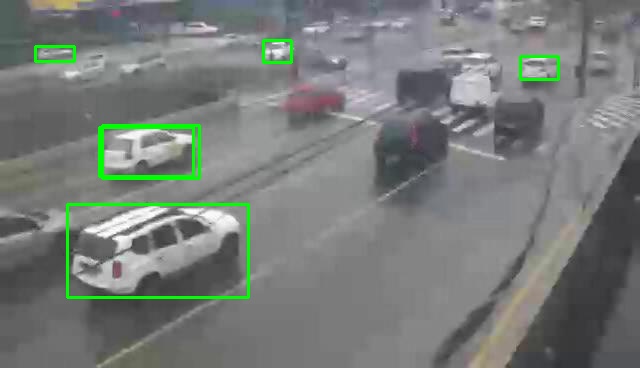}	&
		\includegraphics[width=.14\linewidth]{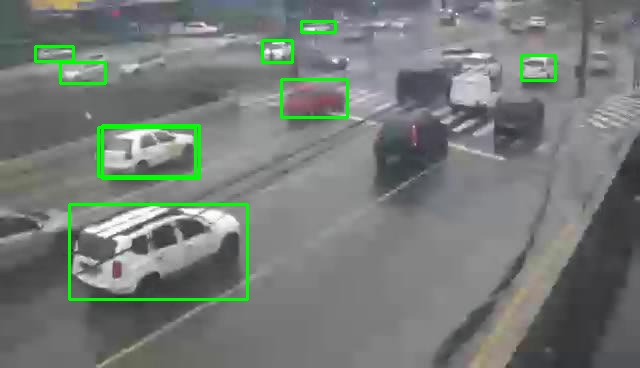}	&
		\includegraphics[width=.14\linewidth]{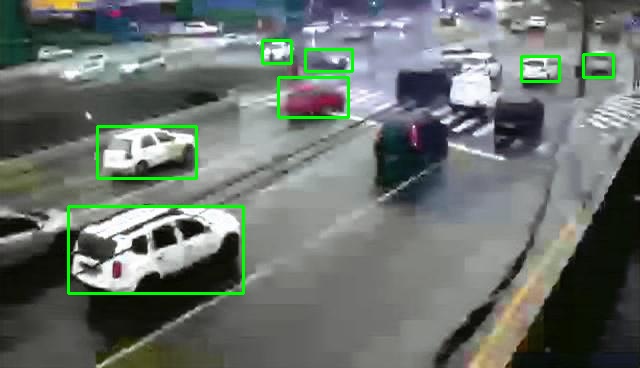}	&
		\includegraphics[width=.14\linewidth]{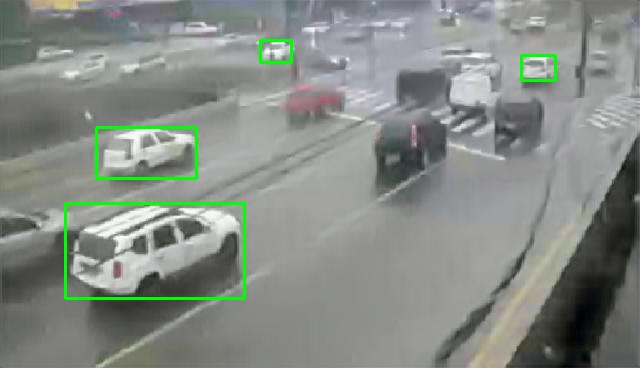}	&
        \includegraphics[width=.14\linewidth]{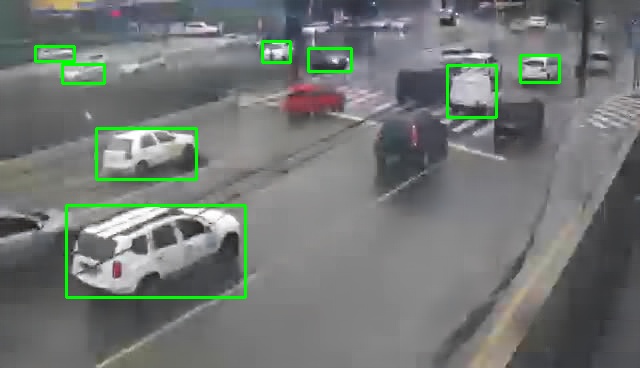}	&
		\includegraphics[width=.14\linewidth]{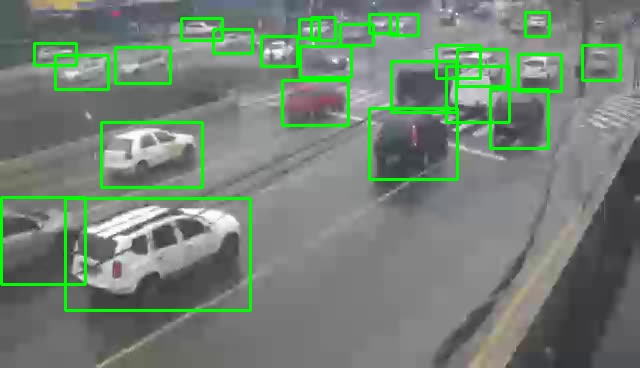}	\\
		(a) Rainy input & (b) JORDER \cite{yang2017deep}  & (c) DDN \cite{fu2017removing} & (d) CGAN \cite{zhang2017image}  & (e) DID-MDN \cite{zhang2018density} & (f) DeRaindrop \cite{qian2018attentive} &(g) Ground-truths \\
	\end{tabular}
	\caption{Visualization of object detection results after applying different deraining algorithms on two images (first two rows) from the RID dataset and two examples (last two rows) from the RIS dataset.
}
\vspace{-5mm}
	\label{fig:detection}
\end{figure*}

\vspace{-1mm}
\subsection{Task-driven Comparison}
\vspace{-1mm}
We first apply all deraining algorithms except GMM\footnote{We did not include GMM for the two sets, because (1) it did not yield promising results when we tried to apply it to (part of) the two sets; (2) it runs very slow, given we have two large sets.}, to pre-processing the two task-driven testing sets. Due to their different rain characteristics, for the RID set, we use deraining algorithms trained on the \textit{rain and mist} case; for the RIS set, we use deraining algorithms trained on the \textit{raindrop} case. We visually inspected the derained results and found the rains to be visually attenuated after applying the selected deraining algorithms.
We show some derained results on the RID and RIS sets in the supplementary material.

We then study object detection performance on the derained sets, using several state-of-the-art object detection models: Faster R-CNN (FRCNN)~\cite{NIPS2015_5638}, YOLO-V3~\cite{redmon2018yolov3}, SSD-512~\cite{liu2016ssd}, and RetinaNet~\cite{lin2018focal}. Finally, we compare all deraining algorithms via the mean Average Precision (mAP) results achieved. It is important to note that our primary goal is not to
optimize detection performance in rainy days, but to use a strong
detection model as a fixed, fair metric on comparing deraining
performance from a complementary perspective. In this way, the object detectors should not be adapted for rainy or derained images, and we use all authors' pre-trained models on MS COCO.
The underlying hypothesis is: i) an object detector trained on clean natural images will perform the best, when the input is also from the clean image domain or close; ii) for detection in rain, the better the rain is removed, the better an object detection model (trained on clean images) will then perform. Such task-specific evaluation philosophy follows \cite{kupyn2017deblurgan,li2019benchmarking}.

%
%

Table \ref{tab-det-RID} reports the mAP
results comparison for different deraining algorithms, achieved using four different detection models, on both RID and RIS sets. We find that quite aligned conclusions could be drawn from the two sets.

Perhaps surprisingly at the first glance, we find that almost \textbf{all existing deraining algorithms will deteriorate the detection performance compared to directly using the rainy images}\footnote{The only exception is FRCNN on the RID set. However, its overall mAP result is the worst compared to the other three. That implies a strong domain mismatch, suggesting that FRCNN results might not be as reliable an indicator for RID deraining performance as the other three.}, for YOLO-V3, SSD-512, and RetinaNet. Our observation concurs the conclusion of another recent study (on dehazing) \cite{pei2018does}: since those deraining algorithms were not trained/optimized towards the end goal of object detection, they are unnecessary to help this goal, and the deraining process itself might have lost discriminative, semantically meaningful true information.

Both results on RID and RIS sets in Table \ref{tab-det-RID} show that YOLO-V3 achieves best detection performance, independently of deraining algorithms applied. Figure~\ref{fig:detection} shows detections using YOLO-V3 on the respectives rainy images and their derained results for all deraining algorithms considered in this comparison. Since both RID and RIS have many small objects due to their relative long distance from the camera, we believe that here YOLO-V3 benefits from its new multi-scale prediction structure, that is known to improve small object detection dramatically \cite{redmon2018yolov3}. We further notice a fairly weak correlation between the mAP results with the no-reference evaluation results of the derained images: see supplementary for more details.




\section{Conclusions and Future Work}
This paper proposes a new large-scale benchmark and presents a thorough survey
of state-of-the-art single image deraining methods. Based on our evaluation and analysis, we present overall remarks and hypotheses below, which we hope can shed some light on future deraining research:
\begin{itemize}
\vspace{-2mm}
\item Rain types are diverse and call for specialized models. Certain models or components are revealed to be promising for specific rain types, e.g., rain detection /attention, GANs, and priors like patch-level GMM. We also advocate a combination of appropriate priors and data-driven methods.
\vspace{-2mm}
\item There is no single best deraining algorithm for all rain types. To deal with the real complicated, varying rains, one might need consider a mixture model of experts. Another practically useful direction is to develop scene-specific deraining, e.g., for traffic views.
\vspace{-6mm}
\item There is also no single best deraining algorithm under all metrics. When designing a deraining algorithm, one needs be clear about its end purpose. Moreover, classical perceptual metrics themselves might be problematic to evaluate deraining. Developing new metrics could be as important as new algorithms.
\vspace{-2mm}
\item Algorithms trained on synthetic paired data may generalize poorly to real data, especially on complicated rain types such as rain and mist. Unpaired training \cite{zhu2017unpaired} on all real data could be interesting to explore.
\vspace{-2mm}
\item No existing deraining method seems to directly help detection. That may encourage the community to develop new robust algorithms
to account for high-level vision problems on real-world rainy images. On the other hand, to realize the goal of robust detection in rain does not have to adopt a de-raining pre-processing; there are other domain adaptation type options, e.g., \cite{chen2018domain}, which we will discuss in future work.
\end{itemize}



{\small
{\flushleft \textbf{Acknowledgments.}}
This work is supported in part by the National Natural Science Foundation of China
(No. 61802403) and CCF-DiDi GAIA (YF20180101). The work of Z. Wang is supported in part by
the US National Science Foundation under Grant 1755701.
}

{\small
\bibliographystyle{unsrt}
\bibliography{egbib}
}

\end{document}